\documentclass[letterpaper, 10 pt, conference]{ieeeconf}  % Comment this line out if you need a4paper  % Comment this line out if you need a4paper

\IEEEoverridecommandlockouts                              % This command is only needed if 
\usepackage{xcolor}
\usepackage{float}
\usepackage{wrapfig}
\usepackage{amsmath}
\usepackage{booktabs}
\usepackage{multirow}
\usepackage{diagbox}
\usepackage{graphicx}
\usepackage{url}
\usepackage{caption}
\usepackage{subcaption}
\usepackage[ruled,vlined,noend,titlenotnumbered]{algorithm2e}
\usepackage{bbm}
\usepackage{amssymb}
\usepackage{siunitx}
\usepackage{wrapfig}
\usepackage{algorithmic}
\usepackage{soul}

\newcommand{\revised}[1]{\textcolor{black}{{#1}}}

\newcommand{\x}{\mathbf{x}}
\newcommand{\f}{\mathbf{f}}

\begin{document}

\title{\LARGE \bf
FastMimic: Model-based Motion Imitation for Agile, Diverse and Generalizable Quadrupedal Locomotion
}

\author{Tianyu Li$^{1}$, Jungdam Won$^{2}$, Sehoon Ha$^{1}$, Akshara Rai$^{2}$% <-this % stops a space
% \thanks{*This work was not supported by any organization}% <-this % stops a space
% \thanks{ Manuscript received: Oct, 15, 2020; Revised: Dec, 17, 2020; Accepted: Feb, 17, 2021.}
% \thanks{This paper was recommended for publication by
% Editor Abderrahmane Kheddar upon evaluation of the Associate Editor and Reviewers’ comments. }
\thanks{$^{1}$Georgia Institute of Technology, Atlanta, GA, USA, 
        {\tt\small \{tli471, sehoonha\}@gatech.edu}}%
\thanks{$^{2}$ Meta AI, USA, 
        {\tt\small \{jungdam, akshararai\}@fb.com}}%
% \thanks{Digital Object Identifier (DOI): see top of this page.}
% \thanks{Extended video: \TL{Add youtube link here}.}
}

\maketitle
% \thispagestyle{empty}
% \pagestyle{empty}

%%%%%%%%%%%%%%%%%%%%%%%%%%%%%%%%%%%%%%%%%%%%%%%%%%%%%%%%%%%%%%%%%%%%%%%%%%%%%%%%
\begin{abstract}

%Learning diverse, agile and generalizable policies for legged robots can be prohibitively expensive in the real world. 
Robots operating in human environments need various skills, like slow and fast walking, turning, side-stepping, and many more. However, building robot controllers that can exhibit such a large range of behaviors is a challenging problem that requires tedious investigation for every task. We present a unified model-based control algorithm for imitating different animal gaits without expensive simulation training or real-world fine-tuning. Our method consists of stance and swing leg controllers using a centroidal dynamics model augmented with online adaptation techniques. We also develop a whole-body trajectory optimization procedure to fix the kinematic infeasibility of the reference animal motions. We demonstrate that our universal data-driven model-based controller can seamlessly imitate various motor skills, including trotting, pacing, turning, and side-stepping. It also shows better tracking capabilities in simulation and the real world against several baselines, including another model-based imitation controller and a learning-based motion imitation technique. 

\end{abstract}

%===============================================================================

\section{Introduction}

Animals are capable of performing diverse and agile locomotion behaviors in nature, but transferring such behaviors on robots remains a challenge. Robots operating in human environments should be equipped with diverse skills like trot and pace when walking slowly or speeding up, turn around corners, and side-step around obstacles. However, building a universal controller that is robust enough to perform such a wide range of skills is still an unsolved problem. 
% \JD{I changed some words so that they matches better to the terminologies used below.}

Motion imitation~\cite{peng2018deepmimic,peng2020learning} holds the promise of learning diverse natural movements by imitating reference motions captured from real humans or animals. Traditional model-based control algorithms~\cite{park2017high,bledt2018cheetah, di2018dynamic} in locomotion literature have demonstrated impressive agility and robustness, but they often require extensive effort in developing the proper mathematical models and control strategies tailored to the given task. On the other hand, many works in physically simulated agent control~\cite{peng2018deepmimic,park2019predict,bergamin2019drecon,won2020scalable,fussell2021supertrack} have demonstrated that motion imitation offers a framework to learn a wide range of skills, including walking, running, jumping, and dancing; it can also easily be combined with existing motion planners such as data-driven kinematic controllers~\cite{holden2017phase}. This approach does not assume any prior knowledge about input motions, and its repertoire is only limited by the given motion dataset and the learning capacity of the policy network, making it ideal for teaching robots a wide range motions. However, learning-based motion imitation is often computationally expensive and its performance could deteriorate during real-world deployment due to sim-to-real gaps unless expensive adaptation processes are performed~\cite{peng2020learning}.

Our key insight is that we can combine the motion imitation paradigm with model-based control to take advantages of both approaches. In the context of model-based control, this data-driven imitation approach provides a unified framework for achieving various motor skills without the burden of task-specific modeling and tuning. From the motion imitation perspective, model-based control can play a role of tracking controllers which do not require training cost per motion unlike learning-based methods. 
% With the help of robustness of model-based control, we can also plug and play many reference motion lively, which is usually not possible for learning-based methods without costly retraining. 
In addition, our controller inherits all the strengths of model-based control, such as better sim-to-real transferability. 

% In this work, we build a model-based controller that is capable of achieving four different animal motions -- trotting, pacing, turning, and side-stepping -- on a quadrupedal robot, starting with animal motion trajectories. Many prior works in locomotion literature, like \cite{bledt2018cheetah, di2018dynamic, peng2020learning} have focused on building dynamic controllers for legged robots. However, very few works demonstrate multiple dynamic gaits using the same underlying controller on legged robots. Developing a universal controller for all the motions is challenging because different motions can have remarkably different characteristics, such as desired center of mass (CoM) motions or foot contact sequences. \cite{peng2020learning} show that reinforcement learning (RL) can be used to transfer animal motions to robots by training one policy per motion. However, RL typically learns a control policy in simulation due to high sample complexity, can result in performance deterioration during real-world deployment. It is customary to fine-tune RL policies on hardware but this fine-tuning can be prohibitively expensive if it needs to be done per motion, per environment. In such cases, a unified controller can not only remove the need for motion-specific policy learning but also make sim-to-real transfer much more efficient. 

\begin{figure}[t]
    \centering
    \begin{subfigure}[b]{0.45\textwidth}
        \centering
        \includegraphics[width=0.95\textwidth]{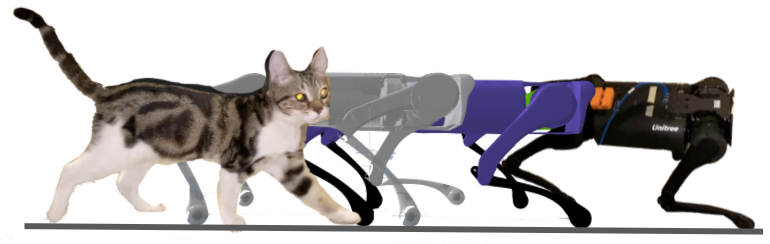}
        \label{fig:A1 Simulation}
    \end{subfigure}
    \label{fig:Profile}
    \caption{\small We present a unified model-based motion imitation algorithm to mimic multiple animal motion trajectories on the A1 robot. We follow animal motions of trotting, pacing, turning, side-stepping using a \textit{single} model-based controller.}
    \vspace{-0.6cm}
    \label{fig:experiment_environment}
\end{figure}

\begin{figure*}[h]
    \centering
    \includegraphics[width=0.85\textwidth]{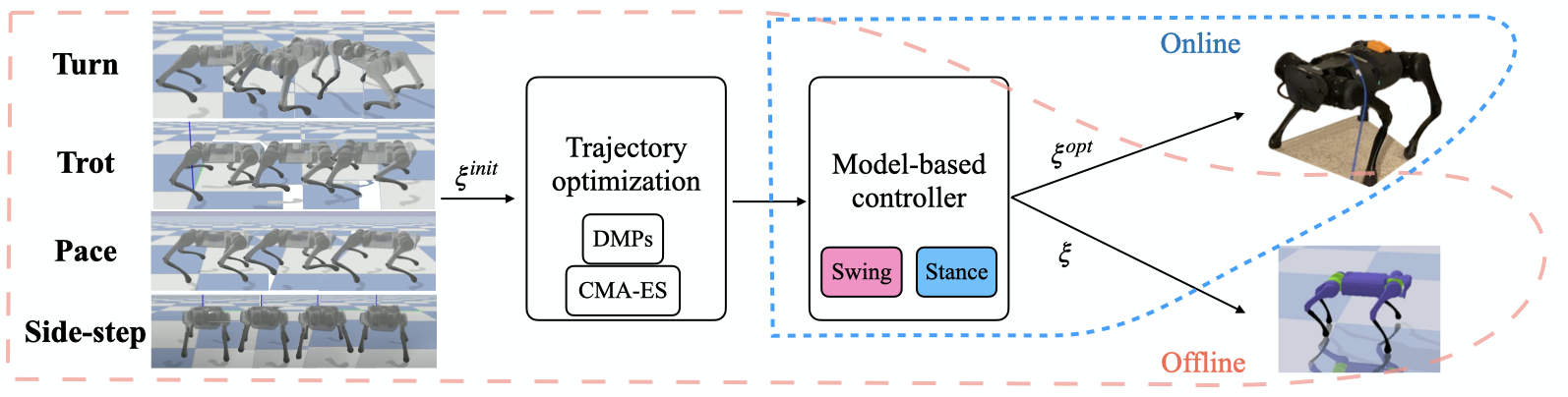}
    \caption{\small An overview of our approach. We start with retargeted animal motions and use them as reference motions in a model-based controller. In simulation, we run trajectory optimization to update the reference motion and send the optimized trajectory to hardware. The optimization pipeline is used on four animal motions - turn, trot, pace, and side-step. The same hardware controller tracks these motions, resulting in diverse, agile gaits on hardware with no real-world fine-tuning. %\sehoon{I think online adaptation box is confusing. I would delete it, and add two small boxes of "stance control" and "swing control" inside of the "model-based controller" box.}  \sehoon{Do we iteratively conduct trajectory optimization and model-based control? If not, I would make the diagram simple.}
    }
    \label{fig:overview}
    \vspace{-0.6cm}
\end{figure*}

% The overall motivation of our work is to combine the power of imitation learning with model-based controllers to result in a universal motion controller, which can track a variety of motions without expensive training. 
We develop a novel model-based control algorithm to imitate a given reference motion. More specifically, our method includes two main components: a stance controller using a low-dimensional centroidal model for the CoM from Di et al.~\cite{di2018dynamic}, along with an online adaptation that commands desired joint positions to improve motion imitation, and a swing controller using inverse-kinematics to follow swing leg trajectories augmented by a feedback term, similar to a Raibert footstep \cite{raibert1986legged}. 
We also develop an algorithm to adapt input animal motions that are kinematically infeasible on target robots, while maintaining the overall style of the motion. We fit a rhythmic dynamic movement primitive (DMP) \cite{ijspeert2013dynamical} to the input motion and optimize a subset of its parameters using CMA-ES \cite{hansen2006cma} to improve imitation performance.
% Our stance controller uses a low-dimensional centroidal model for the CoM from \cite{di2018dynamic}, along with an online adaptation that commands desired joint positions to improve motion imitation. In swing, we use inverse-kinematics to follow swing leg trajectories augmented by a feedback term, similar to a Raibert footstep \cite{raibert1986legged}. The desired CoM motion, swing leg trajectories, and foot contact sequences are extracted from motion capture trajectories of animals and adapted online to improve the stability of the robot. 
% By using different desired trajectories for different motions, we can use the same model-based controller for achieving a wide range of motions. 

% Since the motion capture trajectories from animals can be kinematically infeasible on the robot, we apply motion pre-processing to the raw data and make it kinematically consistent. Next, we fit a rhythmic dynamic movement primitive (DMP) from \cite{ijspeert2013dynamical} to the motion, and optimize part of its parameters using CMA-ES \cite{hansen2006cma} to improve imitation performance. Choosing a DMP parametrization ensures that the trajectory remains close to the original animal motion while CMA-ES optimization makes the trajectory dynamically feasible for the robot. Together, DMP and CMA-ES ensure that the optimized trajectory is well-suited for the robot, including hardware, while staying close to the original animal motion. Once a dynamically feasible and high-performing reference trajectory is found, we transfer the trajectory to hardware without any fine-tuning. 

The contributions of our work are: (1) a universal data-driven model-based controller that can follow any reference motion trajectory and (2) a trajectory optimization algorithm to fix the kinematic infeasibility of the reference motions. We demonstrate that our approach can seamlessly perform various motor skills, including trotting, pacing, turning, and sidestepping, using a \emph{single} controller in simulation and on hardware of an A1 robot~\cite{unitree}. Our controller shows much better tracking capabilities compared to other state-of-the-art model-based imitation control schemes~\cite{kang2021animal} and a model-free imitation learning algorithm~\cite{peng2020learning}.
% The major contributions of our work are: (1) A universal model-based controller and online adaptation scheme that can follow any reference motion trajectory (2) Trajectory optimization that enables the robot to effectively follow different motions, while keeping the `style' of the initial demonstration
% (3) We compare our approach against model-based control with no trajectory optimization, another model-based method from \cite{kang2021animal} and a model-free policy learned using the approach from \cite{peng2020learning}, and demonstrate that our approach outperforms baselines in simulation and on hardware. Our experiments show that a \textit{single} model-based controller, combined with online adaptation and trajectory optimization can generate dynamic, diverse, and natural motions on a quadrupedal robot.

\section{Related Work}

\subsection{Motion Imitation}

\revised{Designing controllers that generate natural-looking motions is challenging on robots and animated characters. Imitating reference motions (usually recorded by motion capture equipment), using learning based approaches has shown a lot of promise in simulation for bipeds~\cite{peng2018deepmimic,bergamin2019drecon,park2019predict,won2019learning,won2020scalable,merel2017learning,merel2019neural,xie2018feedback}, and 
% \cite{peng2018deepmimic,bergamin2019drecon,park2019predict,won2019learning,won2020scalable,merel2017learning,merel2019neural,xie2018feedback,xie2020allsteps}
quadrupeds~\cite{peng2018deepmimic,won2019learning,luo2020carl}. 
In some cases where it is hard or impossible to get those real reference motions, manually created motions could also be used~\cite{peng2018deepmimic, won2019learning}. 
However, learned imitation policies are often vulnerable to the sim-to-real gap even after a long learning process~\cite{xie2018feedback}, which can be mitigated by domain randomization~\cite{exarchos2021policy} or online adaptation~\cite{peng2020learning}. However, fine-tuning per motion in the real world is quite expensive. Instead, we present a unified controller capable of producing varied gaits, without fine-tuning on hardware.}

\subsection{Model-based Legged Locomotion Control}

Model-based approaches use a dynamics model to optimize actions with respect to a given cost/reward function. \revised{Several reduced-order dynamic models have been developed for legged robot control due to its complex dynamics.} For bipedal robots, the inverted pendulum model and its variants have been wildly used~\cite{schwind1998spring,green2021learning,gong2020angular,li2019using}.  For quadrupedal robots, simplifying the robot into a single rigid body that is driven by the sum of external forces from stance legs is \revised{one of the reliable approaches to control~\cite{bledt2018cheetah,kim2019highly, di2018dynamic}. However, these low-dimensional models depending on hand-designed reference motions generate gaits of which styles are robotic and not lifelike. There have been other previous work that combined model-based controllers with reinforcement learning~\cite{xie2021glide,da2020learning} to generate desired CoM acceleration, or adopted learned dynamics models~\cite{9363629} for planning.} 
In this work, we adapt the model-based controller from \cite{di2018dynamic} and use animal motion trajectories as the reference motion to generate diverse, agile, natural motions on an A1 quadrupedal robot. Instead of using RL like \cite{xie2021glide}, we use trajectory optimization to improve the performance of the model-based controller in simulation, and transfer optimized reference trajectory to the real robot.

Closely related to our work, recently Kang et al.~\cite{kang2021animal} present a model-based controller that uses animal motions to create reference CoM trajectories for quadrupedal robots. \revised{However, their swing motion is not inspired from animals, they rather use a Raibert \cite{raibert1986legged} stepping policy with hand-designed foot trajectories. Moreover, all the results are shown in simulation, which include trotting-like walking behaviors only.} In contrast, we present four dynamic gaits -- trot, pace, turn and side-step on an A1 robot in the real-world, using the swing trajectories recorded from animals, resulting in more natural looking motions. To \revised{achieve} this challenging result, we present online adaptation techniques, as well as trajectory optimization to improve reference motion.

\subsection{Dynamic Movement Primitives (DMPs)}
Dynamic Movement Primitives (DMPs) are powerful and well-studied tools for motion imitation. In manipulation problems, DMPs have been widely explored for imitation learning ~\cite{kormushev2010robot,mulling2013learning, ude2010task,conkey2019active}, hierarchical learning ~\cite{kober2009learning,pastor2011skill}, motion optimization ~\cite{rai2017learning, stulp2012path} and reinforcement learning~\cite{bahl2020neural}. In legged robots, pattern generators, a subset of rhythmic DMPs, have been long used for modeling trajectories for locomotion~\cite{jalics1997pattern, kajita2002realtime, yang2004infant}. \revised{However, central pattern generators \cite{ijspeert2008central} and other limit cycle approaches are not easily applicable to imitation learning scenarios. On the other hand, rhythmic DMPs can be learned from demonstrations and applied to locomotion~\cite{7101606}. We learn the parameters of rhythmic DMPs from animal motion, followed by optimizing a subset of the parameters to improve imitation performance.}% on the robot.}

\section{Model-based Control for Motion Imitation}
\begin{figure}
    \centering
    \includegraphics[width=0.35\textwidth]{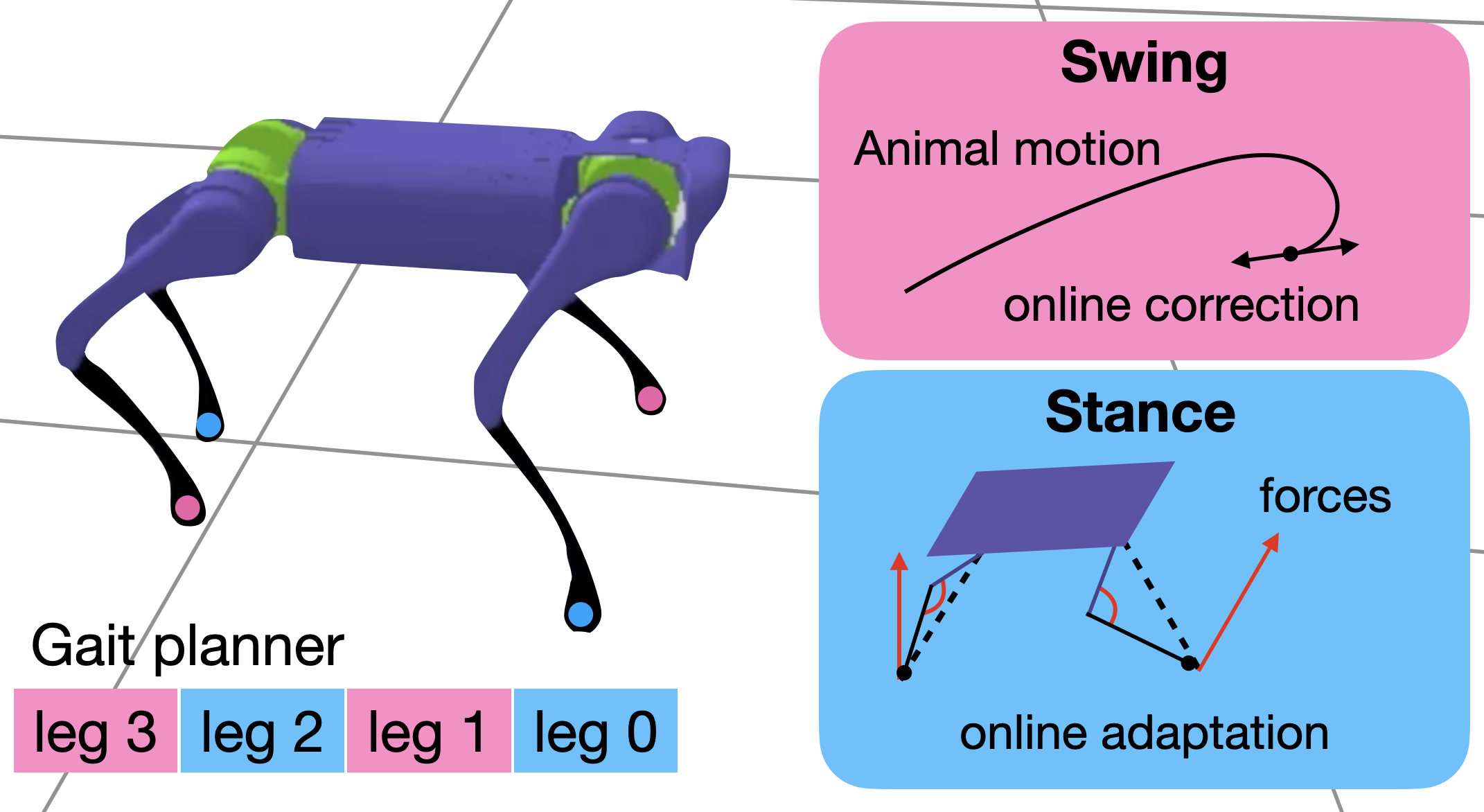}
    \caption{\small An overview of our model-based controller which uses animal motions as desired trajectory input.}
    \label{fig:model-based-overview}
    \vspace{-0.6cm}
\end{figure}

We present a motion imitation framework based on a model-based low-level controller that can generalize to different target motions. \revised{We assume that the reference motions are generated using kinematic motion retargeting with hand-designed keypoints that maps animal motions to the robot.} Then our framework consists of two steps. \revised{(1) Trajectory optimization using a simulated robot equipped with a model-based controller.}
% (1) Trajectory optimization with model-based control in the loop in simulation. %This makes the kinematic trajectory dynamically feasible and well-performing on the robot. 
(2) Transfer of the optimized trajectory to a real-world robot using a model-based low-level controller. The trajectory optimization is performed offline in simulation, and the optimized trajectory is zero-shot transferred to the robot, without fine-tuning. Figure \ref{fig:overview} gives an overview. 

\subsection{\revised{Motion Retargeting}}
\label{retargeting}
We use the retargeting technique described in Peng et al. \cite{peng2020learning} to map animal motions to an A1 robot. We select key-point pairs between animal and robot, followed by inverse-kinematics and manual scaling.
% to map animal motions to the robot. 

%Next, we describe trajectory optimization in simulation and deploy the optimized trajectory on a real-world A1. Both the motion retargeting and trajectory optimization are performed offline in simulation, and the optimized trajectory is zero-shot transferred to the robot, without fine-tuning. 

% In the following sections, we first introduce our model-based controller, followed by an online adaptation technique which improves robustness of the model-based controller. Next, we describe our trajectory optimization approach, using DMPs as the chosen parameterization. Together, this framework leads to a sample-efficient and robust framework for mimicking animal motions on quadrupedal robots.

\subsection{Model-based Controller}
\label{model-based policy}

Our model-based controller consists of three components: (1) A stance controller that commands desired joint angles and torques on the stance legs, (2) a swing controller that commands desired joint angles on the swing legs, and (3) a gait planner switching the legs between swing and stance. The stance and swing controllers take retargeted trajectories as reference and produce outputs based on the current and desired robot states. \revised{The gait planner takes into account the current and desired contact states to determine either swing or stance control per leg. Additionally, we develop an online adaptation scheme to enhance stability by modulating the reference motion according to the current state of the robot.} Figure \ref{fig:model-based-overview} gives an overview of our model-based controller. %We describe the online adaptation for the swing and stance controllers separately below.

\subsubsection{Gait Planner}

\begin{figure}[t]
    \centering
    \includegraphics[width=.4\textwidth]{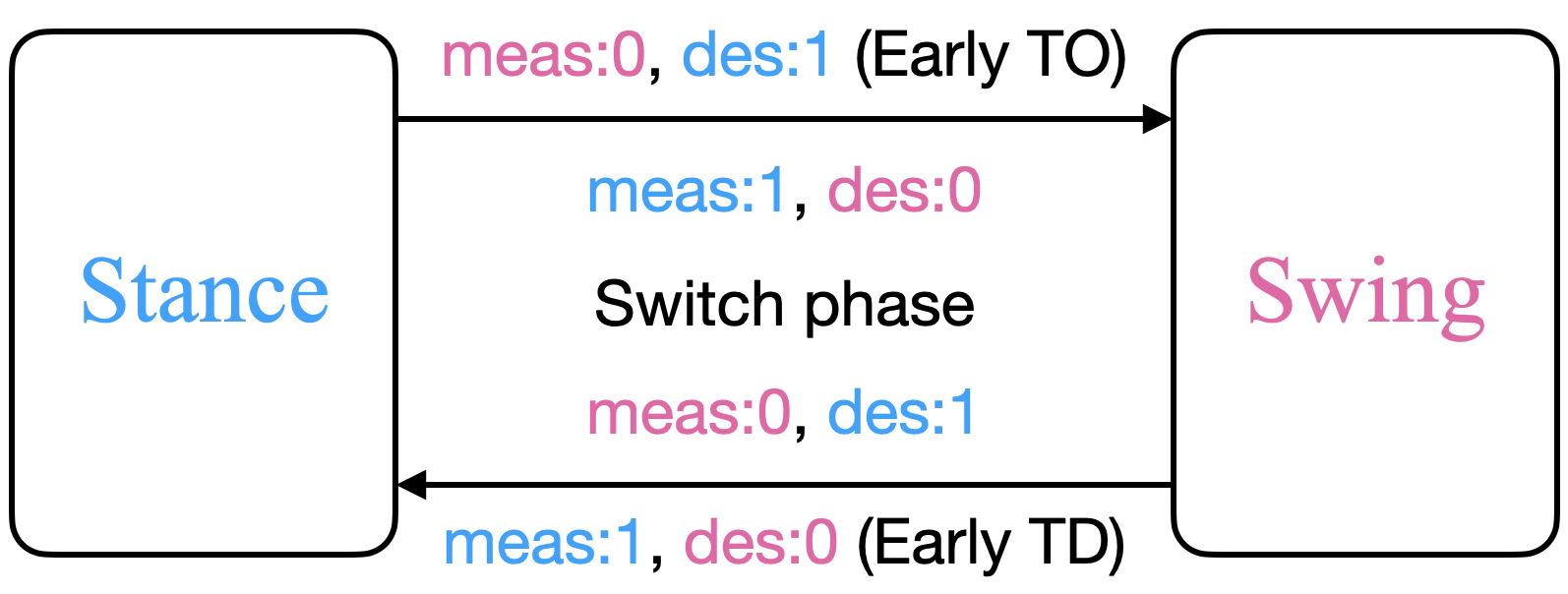}
    \caption{\small Finite state machine used by the gait planner to switch leg control from swing (0) to stance (1) and vice versa. When the measured contact state deviates from the desired contact state, depending on the previous state of the leg, the gait planner initiates a switch from stance to swing, or vice versa. This also takes into account early touch down (TD) and take off (TO). We additionally add a minimum time limit of $100ms$ between consecutive switches to avoid instability due to measurement noise. %\sehoon{Merge two stance states? It looks like when there’s early TD, we can escape the state only with early TO.}
    }
    \label{fig:gait_fsm}
    \vspace{-0.3cm}
\end{figure}

\revised{The gait planner switches the control between swing and stance for each leg, based on the desired contact state, and the current contact state of the robot. The desired contact state is measured from the animal motion demonstration whereas the current contact state is measured by foot sensors on the simulated/real A1 robot. If the measured vertical ground reaction force $\f_{i} > \f_{\text{thresh}}$ for leg $i$, where $\f_{\text{thresh}}$ is the contact threshold, the leg is considered to be in stance.}
\revised{Figure \ref{fig:gait_fsm} shows the finite state machine (i.e. logic) used by our gait planner to determine the appropriate control mode for each leg. When the desired contact state deviates from the measured contact state, the gait planner initiates a transition from swing control to stance control or vice versa. For example, if the desired contact for leg $i$ switches to swing from stance, the measured contact state would remain stance for a short time. The gait planner detects this mismatched contact states and initiates swing on leg $i$. In situations like early contact and early take-off, the current contact state can also deviate from the desired contact state. If the measured force $\f$ on a swing leg is above the contact threshold while the desired state is swing, we detect early contact and start stance control on the leg. Similarly, if a stance leg leaves contact ($\f_{i} < \f_{\text{thresh}}$) when the desired state is in contact, we detect early take-off and switch the leg to swing control. Additionally, we add a minimum time of $100ms$ between consecutive state switches to avoid instability caused by noisy foot force measurements. This results in a robust gait planner, which generalizes across different quadrupedal gaits like trotting, pacing and turning.}

\subsubsection{Stance controller}
The stance controller uses a linearized centroidal dynamics model to reason about robot CoM motion, similar to \cite{di2018dynamic}:  $ \ddot{\x} = \mathbf{{M}^{-1}}\f - \widetilde{g},$
where $\mathbf{{M}^{-1}} $ is the inverse inertia matrix of the robot (calculated from the robot's model), $\f$ is the measured contact force on the stance legs, and $\widetilde{g}$ is the gravity vector. We use $\ddot{\x}$ to denote the 6-dimensional CoM acceleration, consisting of both position and orientation terms in the world frame.

First, we calculate a desired CoM acceleration $\ddot{\bar{\x}}$ as  $\ddot{\bar{\x}} = k_p(\bar{\x} - \x) + k_d(\dot{\bar{\x}} - \dot{\x})$, where $\bar{\x}, \dot{\bar{\x}}$ are the desired CoM position and velocity from the reference motion that the robot is following, respectively, $\x, \dot{\x}$ are the measured CoM position and velocity, and $k_p$, $k_d$ are feedback gains, which are tuned in advance and kept fixed for all reference motions.  
Next, we formulate a Quadratic Program (QP) to solve for the desired instantaneous contact forces $\bar{\f}$ that can achieve $\ddot{\bar{\x}}$ subject to friction constraints, and the current contact state of the robot:
\begin{equation}
\label{eq:mbc}
    \begin{aligned}
    \bar{\f} = \arg\min_{\f} \  \Vert  \mathbf{{M^{-1}}}\f - \widetilde{g} - \ddot{\bar{\x}} \Vert_Q + \Vert \f \Vert_R \\
    \text{s.t.} \  f_{z,i} \geqslant f_{z, min}, \ \mathrm{if} \  \text{Stance} , \quad
    f_{z,i} = 0,  \mathrm{if}\ \text{Swing} \\
    -\mu f_{z,i} \leqslant f_{x,i} \leqslant \mu f_{z,i}, \quad
    -\mu f_{z,i} \leqslant f_{y,i} \leqslant \mu f_{z,i},\\
    \end{aligned}
\end{equation}
% \JD{I might be confused with the notations but why don't we double check the equation above? (1) We do not have the definition $\Bar{M}}$, this might be a typo of the inverse mass matrix? (2) If z is the normal force direction, shouldn't the friction coefficient $\mu$ be multiplied by $f_z$ instead of $f_x$ or $f_z$? It might be a typo of $\mu^{-1}$?}

where, $Q, R$ are weight matrices, kept fixed for all motions. $\f_i = [f_{x,i}, f_{y,i}, f_{z,i}]$ is the contact force on leg $i= \{0, 1, 2, 3 \}$ in world coordinates. $\mu$ is an assumed friction coefficient, and the last two constraints ensure that the robot does not violate a friction cone when in contact with the ground. $f_z$ is zero for legs in swing, and higher than a contact threshold for legs in stance. The QP returns desired contact forces $\bar{\f}$ that are converted to desired leg torques for stance legs using the leg Jacobian: $\mathbf{\tau}_i = J_i^T \cdot \mathbf{R}^T \cdot \bar{\f}_i$, where  $\mathbf{R}$ is the rotation matrix that transforms from body to world.

\noindent \textbf{Online adaptation:} The above model-based controller is borrowed from \cite{di2018dynamic}. During our experiments, we observed that this controller does not robustly track a wide range of target motions, especially when the target motion is dynamic or violates linearized centroidal model assumptions, which could happen frequently for motions obtained from real animals. In such cases, the controller parameters need to be tuned per reference motion to avoid instabilities. This require expert intervention and is not scalable to large varieties of motions. Instead, we augment the stance controller with online adaptation to improve the overall robustness of the model-based controller, allowing the \textit{same} parameters to generalize to different motions.

A common failure case for the robot is caused by poor CoM tracking. For example, if the CoM height is lower than desired, the swing legs might hit the ground early, causing instability. However, `tolerance' to CoM tracking error is highly motion dependant, making feedback gains on CoM acceleration hard to design. To improve CoM height tracking across all motions, without requiring motion-specific tuning, we add low-gain joint position feedback to the stance legs. Specifically, at each time step $t$, we first estimate the foot position of stance leg $i$ at the next time step $t+1$, given the desired CoM velocity:
\begin{equation}
\label{eq:foot-adaptation}
    \hat{\x}^{f}_{t+1, i} = \x^{f}_{t, i} - \dot{\bar{\x}}^{\text{robot}}_t \cdot \delta t.
\end{equation}
Here, $\x^{f}_{t, i}$ is the stance foot location of the $i$-th leg at time $t$ in the robot frame, $\dot{\bar{\x}}^{\text{robot}}_{t}$ is the desired CoM velocity in the robot frame, $\hat{\x}^{f}_{t+1, i}$ is the estimated foot location at the next time, and $\delta t$ is the control frequency. In the robot frame, the stance foot moves with inverse of the local CoM velocity, hence the velocity of the stance foot is $- \dot{\bar{\x}}^{\text{robot}}_{t}$; euler integration results in Eq.~\ref{eq:foot-adaptation}.
%
% The robot was not able to track CoM motion well for dynamic gaits, requiring fine-tuning of feedback gains per motion, and instability caused by poor tracking. For example, if the robot cannot track CoM height, swing tracking is affected, resulting in the robot falling over. 
% To enhance stability, we add a low-gain position feedback to stance leg control, where the desired joint angle is calculated using an online adaptation technique. We calculate the desired stance foot position for leg $i$ in the robot frame at the next time step $t+1$:
% \begin{equation}
% \label{eq:foot-adaptation}
%     \hat{\x}^{f}_{t+1, i} = \x^{f}_{t, i} - \dot{\bar{\x}}^{\text{robot}}_t \cdot dt.
% \end{equation}
% Here, $\x^{f}_{t, i}$ is the leg $i$'s stance foot location at time instant $t$,  $\dot{\bar{\x}}_{t}$ is the desired CoM velocity in the robot frame, $\hat{\x}^{f}_{t+1, i}$ is the estimated foot location at the next time instant, and $dt$ is the control frequency.  
% In the robot frame, the stance foot moves with inverse of the local CoM velocity, hence the velocity of the stance foot is $- \dot{\bar{\x}}^{\text{robot}}_{t}$; euler integration results in Eq.~\ref{eq:foot-adaptation}.
%
%
Next, we use inverse kinematics to find the corresponding joint angles per stance leg, and use low-gain position feedback on these desired positions. 

Intuitively, Eq.~\ref{eq:foot-adaptation} estimates the future joint angles that the robot would end up in, if it followed the desired CoM velocity at the current time step. An alternative approach for determining desired joint angles for stance legs could be to use the joint angles from the reference motion directly. However, these might not be feasible on the robot at execution time. For example, if the foot configuration at the start of stance is different from the reference due to early touch down, leg angles throughout the stance phase will be different from reference motion. Accurately following the joint angles from the reference would require a leg to break contact and re-position the foot. Instead, by using Eq.~\ref{eq:foot-adaptation} we can follow the CoM trajectory in stance without requiring very close correspondence between the reference and actual leg joint angles in stance. This online adaptation improves the robustness of the model-based controller, and allows it to follow different animal motions using the same parameters.

\subsubsection{Swing Controller}\label{swing control}

\revised{Swing foot trajectory is an important part determining the motion `style' and has been shown to be distinct for different animal gaits \cite{schaller2009intertarsal}. Hence, following the swing foot motion as close as possible is crucial for realistic motion imitation of different animal gaits.} 
% In swing, we aim to follow the foot trajectory in the animal reference motion as close as possible. 
  
%
%
\noindent \textbf{Online adaptation:} \revised{The reference motion trajectory basically does not change when the robot operates. However, feedback during swing foot placement can be important for disturbance rejection and increasing stability of legged robots \cite{wu20133}. Therefore, we also add online adaptation to the swing foot position to increases robot stability, especially against real-world disturbances like early touch down, while mimicking the original style of the animal motion as close as possible.}

The target foot position ${\x}^{f, ref}_{i}$ of swing leg $i$ consists of the reference motion $\bar{\x}^f_{i}$ and a feedback on the desired CoM velocity $\dot{\bar{\x}}^{\text{robot}}$ in the robot frame:
\begin{equation}
\label{eq:swing-foot}
    {\x}^{f, ref}_{i} = \bar{\x}^f_{i} - K(\dot{\bar{\x}}^{\text{robot}} - \dot{\x}^{\text{robot}})
\end{equation}
% \JD{No definition on $K$}
Here, K is a constant gain term. The feedback term resembles a Raibert \cite{raibert1986legged} stepping policy, and adds a disturbance rejection mechanism to the purely feed-forward animal reference motion. For example, if the CoM velocity goes higher that in the demonstration due to a disturbance, the swing trajectory compensates by stepping outwards and stabilizing the robot, while maintaining the overall style of the animal motion through $\bar{\x}^f_{i}$. Finally, we use inverse kinematics per leg to convert the desired foot position from robot frame to joint angles and follow this joint trajectory using position control.

\subsection{Trajectory optimization with DMPs}
Reference trajectories from animal motions might not be dynamically feasible for the robot, or might violate some assumptions of our model-based controller. For example, our controller assumes that the stance foot does not slip, but some animal motions might include feet sliding on the ground, making them infeasible for our model. Additionally, naively mimicking animal motion might not result in high-performing motion, due to dynamical mismatch between robots and animals. \revised{Changing (i.e. optimizing) input reference motions in a way to resolve this inconsistency would provide better robot performance. The question arises: what is an ideal way to represent input motions and how can we modify them efficiently while maintaining the style of the original demonstration?}

% We run whole-body trajectory optimization with the model-based controller in the loop to improve motion imitation.

%We run trajectory optimization on the reference trajectory to improve the performance of the model-based controller at imitating animal motions. Our choice of trajectory parametrization are dynamic movement primitives (DMPs).

We choose a dynamic movement primitive (DMP) parametrization to represent swing trajectories. DMPs are trajectory generators that combine linear fixed-point attractors with function approximators whose parameters can be learned from demonstrations. We refer to Ijspeert et al.~\cite{ijspeert2013dynamical} for an overview on DMPs; a demonstration $\xi$, which comes from the retargeted animal motion in our case, is represented by learning parameters $(w^1, w^2, \cdots, g, a)$: $ \textit{DMP}(w^1, w^2, \cdots, g, a) \rightarrow \xi $. Weights $\mathbf{w} = (w^1, w^2, \dots)$ are weights of the non-linear function approximator and encode the overall style of the motion, while $g$ represents the start point of the DMP and $a$ is its amplitude. Specifically, we use rhythmic DMPs which encode cyclic motions whose time period, amplitude, and start point can be modulated by changing the parameters of the DMP. We learn separate \textit{DMP}$_i$ for each dimension of the reference trajectory $\xi_i$, using 100 basis functions per DMP. Reference motions include the CoM position in 6-dimensions, CoM linear and angular velocity, and swing foot trajectory in 3-dimensions for 4 legs (a total of 24 DMPs). DMPs are initialized to mimic the animal motion, and then sent as reference motion to a model-based controller in simulation. 

Once DMPs are learned, their parameters are optimized to improve robot performance. More specifically, we maximize the cumulative reward of the resultant trajectory in simulation (reward described in Section \ref{sec:reward}). Note that we optimize $g, a$ only while keeping $\mathbf{w}$ fixed to maintain the original style of the animal motion during trajectory optimization. Figure \ref{fig:tracking_comparison-x} shows an example of a swing foot trajectory modified by the optimization. We use a gradient-free optimization method CMA-ES \cite{hansen2006cma} to optimize the $g, a$ of the DMPs that represent the z-motion of swing trajectories of all 4 legs. This leads to an 8-dimensional optimization where the swing retraction ($g$), and amplitude ($a$) is optimized to maximize episodic reward: $g_{i=1 \cdots 8}, a_{i=1 \cdots 8} = \arg \min_{g_i, a_i} \sum_{t=1}^T r_t$, 
% \begin{equation}
%     g_{i=1 \cdots 8}, a_{i=1 \cdots 8} = \arg \min_{g_i, a_i} \sum_{t=1}^T r_t,
% \end{equation}
where $T$ is the total length of an episode, and $r_t$ is the step reward, detailed in Section \ref{sec:reward}. We optimize the DMP parameters over 200 iterations of CMA-ES taking about 50 minutes in simulation. 

% The optimization is done offline, and the optimized reference motion is executed on hardware using the low-level controller described in previous sections.  

% \JD{Previously, explanations on DMP parameter learning and its parameter optimization were mixed. I separated them out to make them clear. Feel free to adjust if necessary.}

Because our offline whole-body trajectory optimization procedure takes the full dynamics of the robot, including contacts, into account to improve the reference motion, the optimized trajectory can successfully be transferred to hardware without any fine-tuning. Including domain randomization techniques can further improve the robustness of the optimized trajectory, but we leave this for future work.

\section{experiment}

\begin{figure*}[h]
    \centering
    % \begin{subfigure}[b]{\textwidth}
        % \centering
        \includegraphics[width=0.95\textwidth]{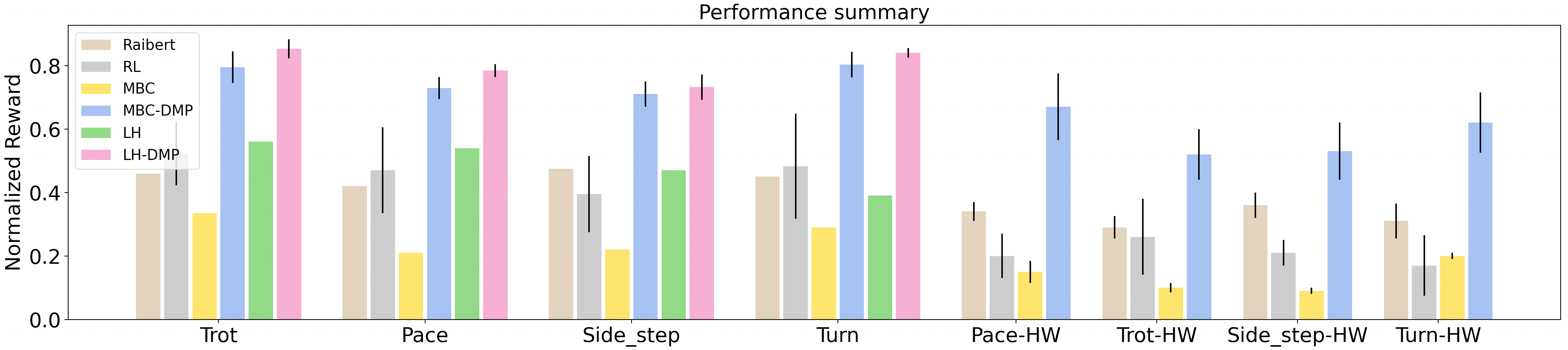}
        \vspace{-0.1cm}
    % \end{subfigure}
    \caption{\small 
    \revised{Comparison against other baselines. Our method (MBC-DMP) outperforms other learning-based (RL) or model-based (Raibert, RL, MBC, LH) in simulation and on hardware, except for LH-DMP. Note that methods with long-horizon predictions, LH and LH-DMP, cannot easily be deployed to hardware due to their computational costs.}
    % Performance of each method. For RL baseline, we train each motion with 2 random seeds and 100 million data samples. For our approach (MBC-DMP) we run optimization with 8 random seeds for each motion, the result is the average of all seeds. Raibert and MBC are deterministic in simulation. For hardware, we take each method's best performing controller in simulation, and run 3 trials on hardware, then average the result. The hardware reward is measured using the robot's IMU data and motor sensors. \sehoon{Remove some experimental details. Focus on the key message (our is better than others. LH and LH-DMP cannot be deployed to the hardware)}
    % \TL{fix the names of the approaches to match text (MBC-DMP instead of DMP-MBC)}
    }
    \label{fig:result-plot}
    \vspace{-0.5cm}
\end{figure*}

We evaluate our method by imitating four animal motions -- trot, pace, turn, and side-step. The motion data, found publicly from Zhang et al.~\cite{zhang2018mode}, is pre-processed using the retargeting procedure (Section~\ref{retargeting}) to map the animal motion to an A1 robot. The target motions can be seen in Figure \ref{fig:overview} and the supplemental video. Our experiments are conducted in a PyBullet simulation~\cite{pybullet} and on hardware, with an A1 quadruped from Unitree Robotics~\cite{unitree} (Fig.~\ref{fig:overview}). A1 is a four-legged robot with 3 servos on each leg. It has 4 force sensors placed on each foot that can measure ground reaction forces, along with joint position, velocity, torque sensors. Robot position is estimated using onboard IMU and kinematic filtering, adapted from open-source code~\cite{peng2020learning}. 

Our experiments show that our method can mimic the desired animal motions in simulation and the optimized trajectory can be zero-shot transferred to hardware. We also present comparisons to RL-based approaches from literature, which learn one policy per motion, and compare performance in both simulation and on hardware. Additionally, we compare against variants of model-based control~\cite{kang2021animal} without trajectory optimization. These experiments help to understand how the different components in our method contribute to the robustness that we achieved.

\subsection{Reward Function}
\label{sec:reward}
We use the same reward function as Peng et al.~\cite{peng2020learning} when training baseline RL policies and measuring overall performance. The per-step reward $r_t \in [0, 1]$ is 
\vspace{-0.1cm}
\begin{align}
    r_t = w^{p}r^{p}_{t}+ w^{v}r^{v}_{t} + w^{e}r^{e}_{t} + w^{rp}r^{rp}_{t} + w^{rv}r^{rv}_{t} \\
[w^{p}, w^{v}, w^{e},  w^{rp}, w^{rv}]= [0.25, 0.05, 0.1, 0.3, 0.3]. \nonumber
\end{align}
Individual components are: 
\begin{itemize}
\item Joint pose reward $r^{p}_{t} = exp [-5 \sum_{j} \Vert  \bar{\mathbf{q}}^j_t - \mathbf{q}^j_t  \Vert^2]$, 
\item Joint velocity reward $r^{v}_{t} = exp [-0.1 \sum_{j} \Vert  \dot{\bar{\mathbf{q}}}^j_t - \dot{\mathbf{q}}^j_t  \Vert^2]$, 
\item End-effector reward $r^{e}_{t} = exp [-40 \sum_{i}\Vert  \bar{\mathbf{x}}^{f}_{t,i} - \mathbf{x}^{f}_{t,i}  \Vert^2]$, 
\item CoM position reward $r^{rp}_{t} = exp [-20 \Vert  \bar{\mathbf{x}}^{robot}_{pos,t} - \mathbf{x}^{robot}_{pos,t}  \Vert^2 -10 \Vert  \bar{\mathbf{x}}^{robot}_{ori,t} - \mathbf{x}^{robot}_{ori,t}  \Vert^2]$, 
\item CoM velocity reward $r^{rv}_{t} = exp [-2 \Vert  \dot{\bar{\mathbf{x}}}^{robot}_{pos,t} - \dot{\mathbf{x}}^{robot}_{pos,t}  \Vert^2 - \Vert  \dot{\bar{\mathbf{x}}}^{robot}_{ori,t} - \dot{\mathbf{x}}^{robot}_{ori,t} \Vert^2]$.
\end{itemize}
Here,  $\bar{\mathbf{q}}^j_t$ and $\mathbf{q}^j_t$ stand for the desired and current joint positions, $\dot{\bar{\mathbf{q}}}^j_t$ and $\dot{\mathbf{q}}^j_t$ are the desired and current joint velocities, $\mathbf{x}^{f}_{t,i}$ represents the current foot position, ${\mathbf{x}}^{robot}_{pos,t}$ and $\dot{\mathbf{x}}^{robot}_{pos,t}$ denote the CoM linear position and velocity, ${\mathbf{x}}^{robot}_{ori,t}$ and $\dot{\mathbf{x}}^{robot}_{ori,t}$ are the CoM angular position and velocity. The higher the reward, the closer the robot motion is to the demonstration.

\vspace{-0.1cm}
\subsection{Comparison experiments}
Here, we describe the different approaches compared in our paper. The baselines are chosen to highlight the robustness of a unified model-based controller, and the efficacy of our approach \textbf{(MBC-DMP)} at mimicking animal motions. Note that all model-based controller baselines also use the online adaptations described in Section \ref{model-based policy}, which was essential to get robust performance across the 4 motions considered in our experiments.

\begin{itemize}

 \item \textbf{DeepMimic (RL):} We compare our approach against DeepMimic~\cite{peng2018deepmimic}, a learning-based approach that learns an RL policy per reference motion. We use the reward function described in Section \ref{sec:reward}, train each policy for 100 million simulation steps using Proximal Policy Optimization \cite{schulman2017proximal}. We use the open-source implementation to train policies for trot, pace, turn, and side-step motions. For each target motion, we train 2 RL policies with different random seeds and report average performance in Figure \ref{fig:result-plot}. We apply policies learned in simulation to hardware with no fine-tuning, to make the comparison fair to our approach.

\item \textbf{Model-based Controller with Raibert Swing (Raibert):} Next, we compare our method against a model-based method from Kang et al.~\cite{kang2021animal} which uses animal reference for CoM motion, but uses linear swing trajectories of fixed time length $T_s$ that reach a footstep calculated using the Raibert heuristic: ${\x}^{f, ref}_{i} = 0.5 T_s \dot{\x}^{\text{robot}} - K(\dot{\bar{\x}}^{\text{robot}} - \dot{\x}^{\text{robot}})$. 
% \begin{equation}
%     {\x}^{f, ref}_{i} = 0.5 T_s \dot{\x}^{\text{robot}} - K(\dot{\bar{\x}}^{\text{robot}} - \dot{\x}^{\text{robot}})
% \end{equation}
Compared to Eq. \ref{eq:swing-foot}, we note that this method does not take the animal motion into account, while our approach augments the animal motion with a stabilizing feedback mechanism. Because each gait has unique swing foot motion style, using pre-defined swing trajectories could have difficulties in reproducing natural and diverse motion styles. %, making it important for robot motion as well.

\item  \textbf{Model-based Controller (MBC) :} 
In this baseline, we send the animal motion trajectories to our model-based controller without any trajectory optimization. This experiment highlights the importance of whole-body trajectory optimization in simulation.

\item \textbf{Model-based Controller with DMP Optimization (MBC-DMP, ours): }
Our approach which uses offline trajectory optimization to adapt the reference motion sent to the model-based controller.

\item \textbf{Long horizon Model-predictive control (LH): }
Whole-body control of quadrupedal robots can be improved by using a receding-horizon model predictive control approach that plans over multiple time steps, instead of instantaneous forces, as in Eq. \ref{eq:mbc}. We utilize the approach from Di et al.~\cite{di2018dynamic} to solve a higher-order QP to plan actions over a horizon of $10$ steps, instead of a single step. During stance, this long-horizon convex MPC controller uses the linearized centroidal model to predict future states, and plans a sequence of contact forces that lead to a desired CoM trajectory. For swing control, we use the same setup as ours. This experiment is aimed to test if online MPC can replace offline trajectory optimization.

\item \textbf{Long horizon MPC with DMP Optimization (LH-DMP): }
\revised{Lastly, we augment the long horizon model-predictive control (LH) with our DMP-based trajectory optimization (LH-DMP). We expect that this method outperforms our method because it reasons over a longer horizon of the reference motion. However, both long-horizon predictive control methods, LH and LH-DMP, cannot easily be deployed to hardware due to their expensive computational costs. Therefore, we only conduct simulation experiments for analysis.}
% Lastly, we consider a variant of our approach, by replacing the instantaneous optimization in Eq \ref{eq:mbc} with a long-horizon convex MPC controller from \cite{di2018dynamic}. The reference motion sent to the MPC is optimized using trajectory optimization, same as our approach. The long-horizon convex MPC is strictly better than our approach, because it reasons over a longer horizon of the reference motion. However, this controller is slower to execute, and hence challenging to apply to hardware without significant code speed-up. \sehoon{This slow computation is the shared problem for both LH and LH-DMP, no?} Here, we conduct simulation experiments to compare the performance gains of long-horizon over instantaneous optimization in our problem. 
\end{itemize}

\begin{figure*}
    \centering
    \includegraphics[width=0.9\textwidth]{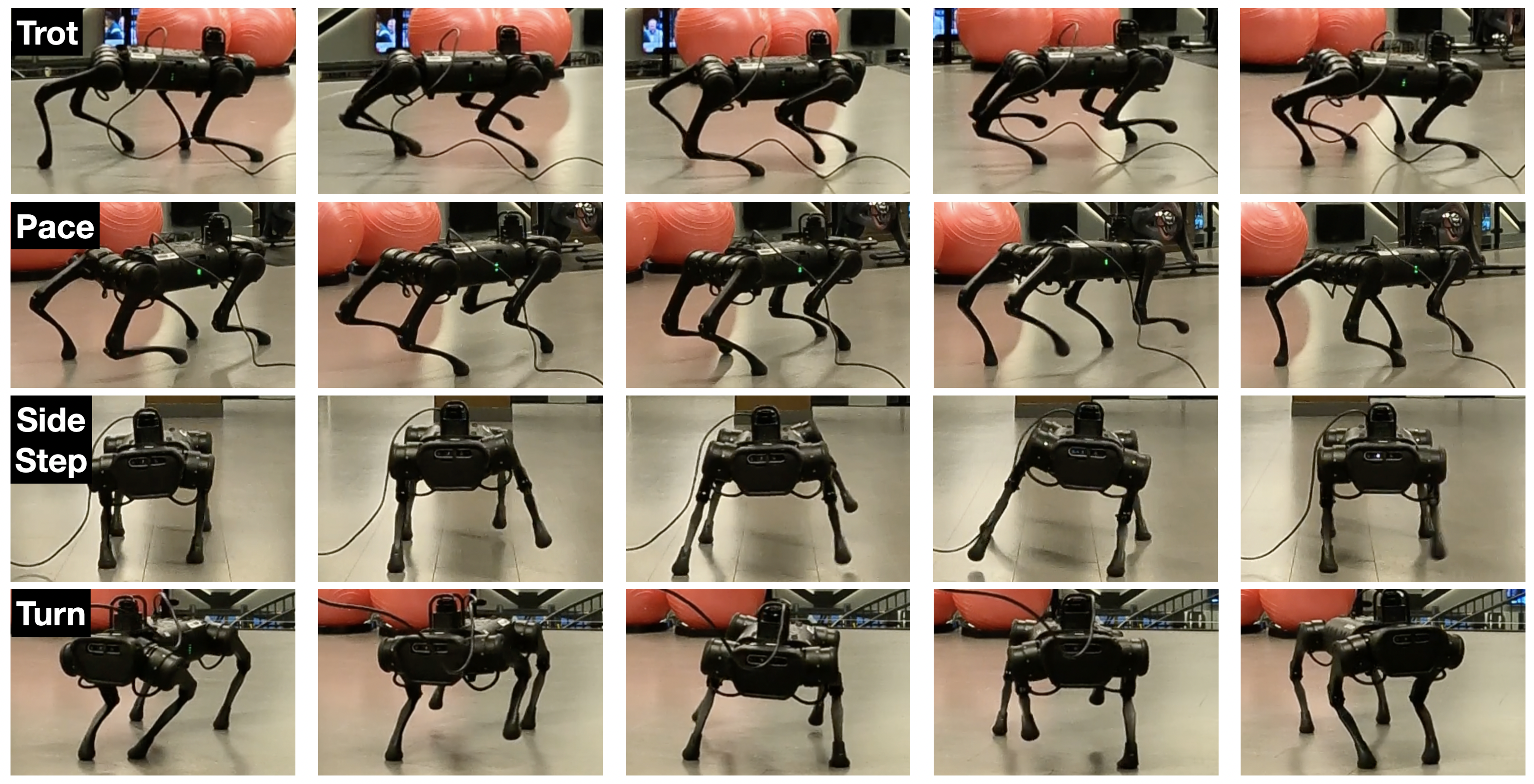}
    \caption{\small Snapshots of our approach in action on the A1 robot - Turn, Trot, Pace and Side-stepping motions. 
    % Blue dots highlight stance feet and pink highlight swing feet. \sehoon{No blue and pink dots now.}\TL{I think this figure the contact/noncontact is obvious, do we still need these dots?}\sehoon{Either way is fine, just be consistent.} 
    A \textit{single} model-based controller with online adaptation achieves all the motions, with no fine-tuning.
    % \TL{can you use a little further apart frames for turn so that the robot looks like it turned?}
    }
    \label{fig:snapshots}
    \vspace{-0.6cm}
\end{figure*}

Figure \ref{fig:result-plot} shows comparison experiments between the different baseline approaches and our approach (MBC-DMP). In simulation, we observe that DeepMimic (RL) is able to replicate all target motions well. However, when transferred to hardware, we see a significant drop in performance, owing to the sim-to-real gap (e.g. $0.483 \pm 0.195$ in sim vs. $0.17 \pm 0.095$ on hardware for turn). \revised{We observe that RL policies tend to be very conservative, and do not lift the robot legs sufficiently during swing. This causes early-contact on hardware which degrades the performance on the real robot. Peng et al.~\cite{peng2020learning} show that online adaptation can improve the performance of RL on hardware. In contrast, our method can generalize to hardware with no real-world adaptation.}

\begin{figure}
% \vspace{-0.1cm}
    \centering
    \begin{subfigure}[b]{0.235\textwidth}
        \centering
        \includegraphics[trim={0.9cm 0 0 0},clip, width=\textwidth]{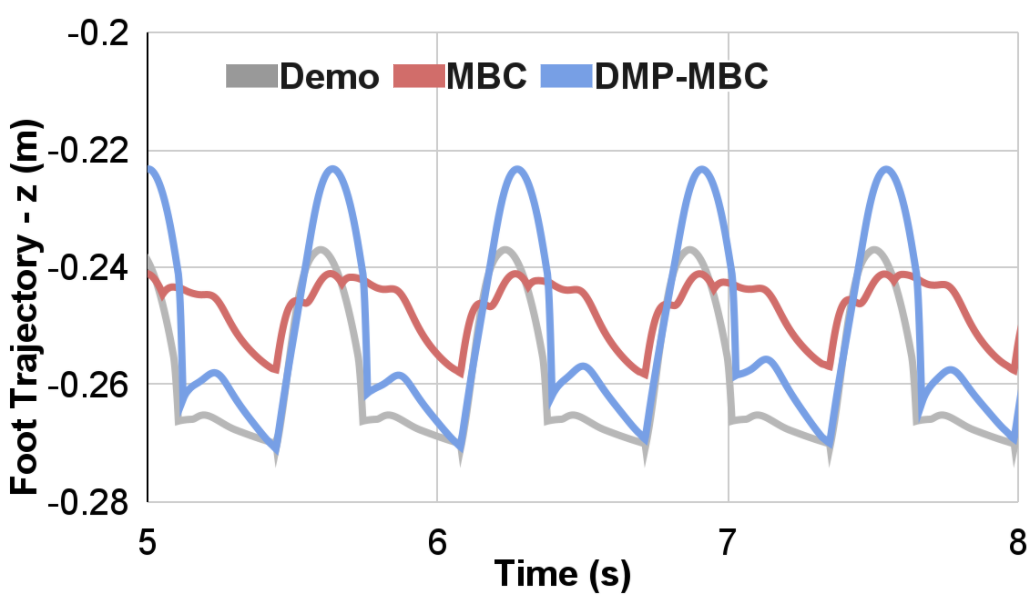}
        \caption{\small Foot position in z-axis (m)}
    \label{fig:opt_comparison-z}
    \end{subfigure}
    \centering
    \begin{subfigure}[b]{0.235\textwidth}
        \centering
        \includegraphics[trim={1.2cm 0 0 0},clip,width=\textwidth]{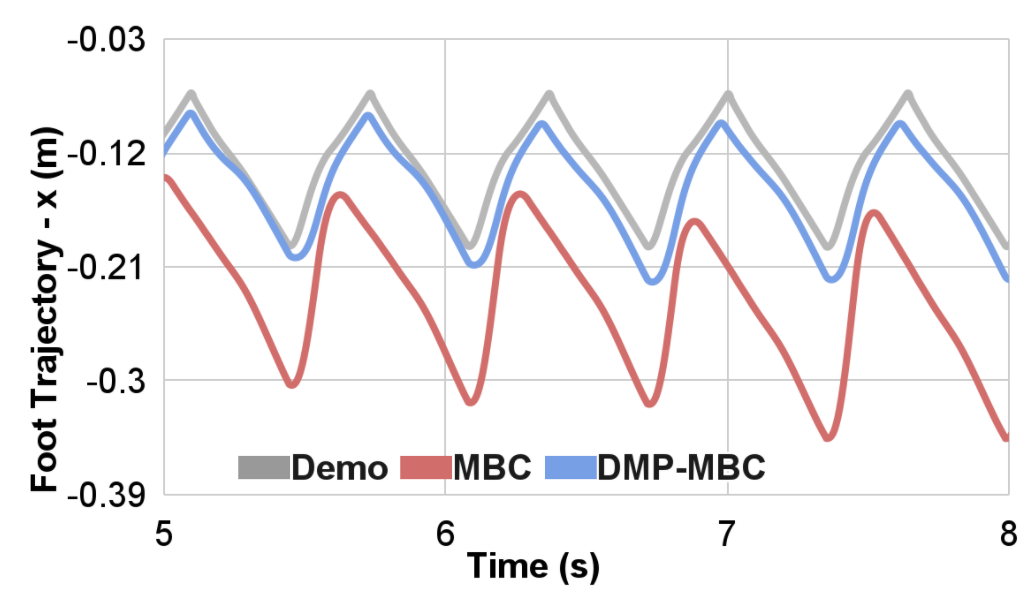}
        \caption{\small Foot position in  x-axis (m)}
    \label{fig:tracking_comparison-x}
    \end{subfigure}
    \caption{ \small (a) Tracking in z-axis. DMP optimization increased the amplitude of the trajectory, tracking error was decreased by 24$\%$. (b) Tracking in x-axis. With optimization, the constant drift was reduced, tracking error was decreased by 92$\%$. 
    % \sehoon{Negative z axis values sound a bit weird to me..}\TL{Here is foot position in the robot frame, thus z is negative}
    }
    \vspace{-0.6cm}
\end{figure}

Model-based control with no trajectory optimization (MBC) achieves good performance in pace and trot, but poor performance in side-step and turn in simulation (e.g. $0.223$ using MBC vs. $0.715 \pm 0.012$ using our approach for turn). In side-step and turn, the robot falls over due to low swing leg retraction, and early contact, increasing the disturbance on the system. When applied to the real robot, the pace and trot motions also deteriorate in performance. As an example, Figure \ref{fig:tracking_comparison-x} shows the tracking of the left back leg during a pace motion in simulation. The x-tracking of the foot deviates from the reference during swing, made worse on hardware due to tracking errors. Although online adaptation of the model-based controller manages to maintain robot stability, the overall imitation reward is very low. On the other hand, our approach learns to optimize the swing leg retraction in simulation, also improving performance on hardware.

The model-based baseline Raibert achieves lower performance than our approach on all gaits, with turning motion seeing the largest drop in simulation ($0.453$ vs. $0.803 \pm 0.008$). Turning is a challenging motion, with robot legs stretching outwards, more in the rear legs and lesser in the front legs. If the swing leg trajectories are designed to be symmetric across all legs, as in \cite{kang2021animal}, not only is the swing motion not natural, but also the CoM turning motion does not follow the reference. \revised{This experiment highlights the importance of following both reference (animal) swing and stance motions to generate natural motions.} Due to the robustness of Raibert swing trajectories combined with our model-based controller, Raibert does not experience a significant performance drop between sim and real, but the overall performance is still lower than our approach.

Our approach (MBC-DMP) outperforms all RL- and model-based baselines that plan instantaneous actions in simulation and on hardware. 
The DMP optimization improves motion tracking in simulation, and also generalizes to hardware with little performance deterioration in most gaits (Fig. \ref{fig:result-plot}). Fig. \ref{fig:opt_comparison-z} compares the optimized vertical foot motion with the original trajectory of the back left leg in a pace gait, where the optimization increases the leg clearance for swing trajectory, improving tracking and stability. 
Figure \ref{fig:snapshots} shows that the optimized DMP reference trajectory achieves good hardware performance on challenging motions like turning ($0.803 \pm 0.012$ in sim vs. $0.627 \pm 0.096$ on hardware). In the side-step motion, we observe deterioration in performance on hardware due to slipping. Side-step requires a wide robot stance which makes the leg configuration at the edge of the friction cone, making it easy for the robot to slip. Since we do not do any system identification on the actual friction of the floor, the model-based controller is unable to compensate for the slippery floor. We believe that this behavior can be improved by real-world fine-tuning, which can be very efficient as once the friction coefficient is identified, it can be used for all motions.

The hypothesis that offline trajectory optimization is not needed if planning over a long horizon is disproved in our experiments. Despite using a longer horizon, and online adaptation, long-horizon MPC (LH) performs worse than our approach on all simulated tasks ($0.39$ using LH vs.$0.803 \pm 0.012$ using our approach on turn). On the other hand, when the reference trajectory sent to long-horizon MPC is optimized using our DMP parametrization (LH-DMP), we see that there is a rise in performance. This further reinforces that even with long horizon planning, offline trajectory optimization is crucial for robust mimicking of varied animal motions. As compared to long-horizon MPC (LH-DMP), we observe that the loss in performance is minor when using instantaneous optimization ($0.831 \pm 0.004$ using LH-DMP versus $0.803 \pm 0.012$ using MBC-DMP on turn.) This motivated our decision to perform hardware experiments using an instantaneous QP, and leave real-world evaluation of costly LH-DMP for future work.

% \subsection{Motion stitching and long-horizon planning} % 
% \revised{newly added section}\\
% In our setting, we use a impulse control as the underlying model-based controller. The impulse control as the advantage of low computational cost which makes it a great candidate of high-frequency hardware control. However, this type of control has the shortage of only focusing on the current state and not taking future target into consideration which makes it not suitable for some challenging tasks. We tackle this issue by simply replacing the impulse control with a convex MPC controller\cite{di2018dynamic}. 

% In this section, we examine how the performance varies as we use a  
% Given the desired CoM trajectory and desired contact state, the MPC controller optimized a sequence of desired contact forces for stance leg to track the desired trajectory. For swing control, we maintain the same control strategy as we proposed in Section \ref{swing control}.

% We compare the 

\subsection{Motion stitching}
We create new motions by stitching different motion trajectories to verify that our method can be used for imitating very long motions. The stitched motions are created by directly concatenating individual reference motions $\xi_{st} = \xi_1 \oplus \xi_2$, for example by concatenating pacing and trotting trajectories. However, these stitched trajectories cannot be directly tracked by the model-based controller since the transition between the motions is not smooth. 
Therefore, we first fit each elementary motion $\xi_{i}$ independently with DMPs. Then, we optimize all DMPs together w.r.t the reward function mentioned in Sec \ref{sec:reward}. To warm start the optimization,  we initialize the DMPs with the parameters for individual optimized motion. 
% Here, the referenced trajectory is $\xi_{st}$.
% To make trajectory optimization faster, we initialize the DMPs with the optimized DMPs from previous experiment.
%
Our method successfully imitates four stitched motions: (1) Pace + Trot, (2) Trot + Turn + Trot, (3) Side Step + Trot + Turn, and (4) Trot + Side Step + Reverse trot. Real-world results can be seen in the supplementary video.

\section{Conclusion}
In this work, we propose a unified framework for transferring agile animal motions to real-world robots. Our framework uses trajectory optimization and a model-based controller with online adaptation. Experiments show that our model-based controller can imitate reference motions robustly, and the quality of imitation can be further improved by trajectory optimization. We validate our framework by applying our model-based controller to four motions in simulation and hardware. We compare our approach with learning-based and model-based baselines. We show that our method outperforms the baseline methods in simulation and on the real-world A1 robot. Although our method currently does not include any sim2real adaptation, incorporating it into our method would bring further capabilities such as enhanced robustness to external perturbation. We are also interested in combining our method with other learning-based methods to learn more dynamics motions, for example, back flip and jumping motions.

%===============================================================================

\bibliographystyle{IEEEtran}
\bibliography{bibliography}

\end{document}